\documentclass[wcp]{jmlr}

\usepackage{longtable}
\usepackage{amsmath}
\usepackage{amsfonts}
\usepackage{multirow}
\usepackage{pgfplots} 
\usepackage{graphicx}
\usepackage{algorithm}
\usepackage[normalem]{ulem}
\usepackage{adjustbox}
\usepackage{float}
\usepackage{rotating}
\usepackage{multirow}
\usepackage{csquotes}
\usepackage{booktabs}
\pgfplotsset{compat=1.18} 
\usepackage{amssymb,amsmath, bm}
\usepackage[hang,flushmargin]{footmisc}
\usepackage{wrapfig}
\usepackage[belowskip=-10pt,aboveskip=2pt]{caption}

\def\comments{} 

\ifdefined\comments
    \definecolor{ENX light blue}{HTML}{00B0F0}
    
    \newcommand{\RemoveForShortVersion}[1]{\textcolor{gray}{#1}}

    \newcommand{\Gianmarco}[1]{\smallskip\noindent{}\textcolor{red}{{\bf G.Av}: #1}}
    \newcommand{\Valentin}[1]{\smallskip\noindent{}\textcolor{violet}{{\bf V.Le}: #1}}
    \newcommand{\Youssef}[1]{\smallskip\noindent{}\textcolor{magenta}{{\bf Y.Ac}: #1}}
    \newcommand{\Houssem}[1]{\smallskip\noindent{}\textcolor{green}{{\bf H.So}: #1}}
    \newcommand{\Lucas}[1]{\smallskip\noindent{}\textcolor{orange}{{\bf L.Od}: #1}}
    \newcommand{\QFE}[1]{\smallskip\noindent{}\textcolor{teal}{{\bf Q. Fe}: #1}}
    \newcommand{\STI}[1]{{\color{ENX light blue}~\{STI: #1\}}}

\else

    \newcommand{\RemoveForShortVersion}[1]{}
    \newcommand{\Gianmarco}[1]{}
    \newcommand{\Valentin}[1]{}
    \newcommand{\Youssef}[1]{}
    \newcommand{\Houssem}[1]{}
    \newcommand{\Lucas}[1]{}
    \newcommand{\QFE}[1]{}
    \newcommand{\STI}[1]{}
\fi
\usepackage{booktabs}

\usepackage{lineno}

\makeatletter
\let\Ginclude@graphics\@org@Ginclude@graphics 
\makeatother

\jmlrvolume{}
\jmlryear{}
\jmlrworkshop{}

\title[SANGEA]{SANGEA: Scalable and Attributed Network Generation}

  \author{\Name{Valentin Lemaire}$^{1^*}$ \Email{valentin.lemaire@euranova.eu}\\
  \Name{Youssef Achenchabe}$^{2^*}$ \Email{youssef.achenchabe@euranova.eu}\\
 \Name{Lucas Ody}$^{1^*}$ \Email{lucas.ody@euranova.eu}\\
 \Name{Houssem Eddine Souid}$^{3^*}$ \Email{houssem.souid@euranova.eu}\\
  \Name{Gianmarco Aversano}$^1$ \Email{gianmarco.aversano@euranova.eu}\\
  \Name{Nicolas Posocco}$^1$ \Email{nicolas.posocco@euranova.eu}\\
  \Name{Sabri Skhiri}$^1$ \Email{sabri.skhiri@euranova.eu}\\
   \addr $^1$ Euranova, Rue Emile Francqui 4, 1435 Mont-Saint-Guibert, Belgium. \\
   $^2$ Euranova, 146 Rue Paradis, 13006 Marseille, France. \\
   $^3$ Euranova, 6 Les Berges du Lac 3, Tunis 2015, Tunisia. \\ $^*$ Equal contributions.}

\begin{document}

\maketitle
\begin{abstract}
The topic of synthetic graph generators (SGGs) has recently received much attention due to the wave of the latest breakthroughs in generative modelling.
However, many state-of-the-art SGGs do not scale well with the graph size.
Indeed, in the generation process, all the possible edges for a fixed number of nodes must often be considered, which scales in $\mathcal{O}(N^2)$, with $N$ being the number of nodes in the graph.
For this reason, many state-of-the-art SGGs are not applicable to large graphs.
In this paper, we present SANGEA, a sizeable synthetic graph generation framework which extends the applicability of any SGG to large graphs.
By first splitting the large graph into communities, SANGEA trains one SGG per community, then links the community graphs back together to create a synthetic large graph.
Our experiments show that the graphs generated by SANGEA have high similarity to the original graph, in terms of both topology and node feature distribution. Additionally, these generated graphs achieve high utility on downstream tasks such as link prediction.
Finally, we provide a privacy assessment of the generated graphs to show that, even though they have excellent utility, they also achieve reasonable privacy scores. 

\end{abstract}


\section{Introduction}

The recent rise in the research community's interest in generative artificial intelligence has also included graph data. 
Graph generation has applications in several domains, from generative chemistry \citep{bian2021generative} for drug discovery to source code generation \citep{brockschmidt2018generative}.
In this paper, we exclusively focus on the setting where, given a single large graph as a training sample, we generate another graph of the same size.
For instance, this happens when a social network is represented as a single graph, making it suitable for studying how people interact and influence each other. In addition, numerous industrial applications where an institution owns a single large graph (mobility data, transactions, calls graphs, etc.) and intends to share it. Generative models are then trained on real data and used to generate synthetic samples to be shared for training models on downstream tasks.

Graphs are represented by their node feature matrix $\mathbf{X} \in \mathbb{R}^{N \times D}$, and by their adjacency matrix $\mathbf{A} \in \{0, 1\}^{N \times N}$, which scales with $\mathcal{O}(N^2)$, with $N$ being the number of nodes in the graph, and $D$ being the number of node features.  
This quadratic complexity makes it very \textit{challenging} to deal with large graphs.
The deep generative learning literature is rich in models that deal with synthetic graph generation \citep{graphrnn, gran, graphnf, graphgen, bigg, nvdiff, gdss}, but most state-of-the-art models still suffer from graphs' intrinsic scalability issues. 
Synthetic graph generators (SGGs) are generally classified in the literature into two main categories: one-shot and recurrent generators.
The former usually requires storing a dense adjacency matrix in memory, which is only feasible for a few nodes.
As for the latter, they take a long time to train because they recursively go through all the nodes in the graph during training and generation. 
Moreover, they are not node-invariant, so the ordering of the nodes matters considerably. 
In addition, since the topology creates dependencies between nodes within a graph, the data parallelisation within a graph is not trivial and often causes overhead.
In summary, graph generation is challenging to scale.

One of the purposes of graph generation is to share data privately.
However, the risks of re-identification still apply to synthetic datasets.
Graphs are not immune to this phenomenon and have actually been shown to leak more private information than other data modalities due to the information they carry in their topology \citep{wu2021adapting}.
For this reason, in the present work, we also provide a privacy assessment methodology by means of nearest neighbour distance ratio (NNDR) \citep{guzman2021unravelling} adapted to graphs.




Our goal in this paper is to generate, from a single large attributed graph, another large attributed graph that matches the statistical properties of the original one while being privacy-preserving.
We present SANGEA (\underline{S}calable and \underline{A}ttributed \underline{N}etwork \underline{GE}ner\underline{A}tion), a lightweight method to scale \textit{any} graph generative models to many nodes and edges under the assumption that the training graph presents a community structure.

The essence of our approach is dividing the input graph into densely connected communities that can be generated independently. Then, SANGEA learns to model inter-community interactions based on independent subgraphs. Since this divide-and-conquer strategy may not leverage joint distributions of the communities and the links between them, SANGEA iteratively improves the generated graph until it matches the original distribution.

SANGEA offers numerous advantages: 
i) it limits the original generation to different graphs with fewer nodes, allowing the use of any high-quality but potentially unscalable state-of-the-art generation method;
ii) only one-shot generation models are used to predict links between communities and to perform the updates, making them fast to learn and fast at inference; 
iii) only node-invariant models are used, making the process more generalizable and less prone to overfitting, which is a challenge as we have only one training sample;
iv) our refinement process conditions the updates on the synthetic graph in a similar manner to recurrent methods, thus removing the need to sample from a high-dimensional joint distribution like other one-shot generation methods do; v) Empirical results show that our proposed method achieves high privacy scores.

The contribution of this paper is threefold. Firstly, it proposes a novel approach to make \textit{any} state-of-the-art model scalable to large graphs that present a community structure. Secondly, extensive experiments are presented on five models from the literature and compare our proposed approach against these models, to show that we match the quality of those other models while allowing us to perform generative model training and sampling for graphs up to 90,000 nodes and 450,000 edges. Thirdly, a privacy assessment has been performed for our generated data.

The rest of this paper is organized as follows. The next section presents essential works related to deep synthetic graph generators. Section \ref{sec:SANGEA-model} details our proposed model by explaining our training and generation procedures. Then, Section \ref{sec:exps} presents the experimental setup and reports results with analyses. Section \ref{sec:conclusion} concludes by highlighting the main findings of this research and by discussing directions for future works.

\section{Related Works}

Many approaches have been considered in synthetic graph generation. On one hand, traditional statistical methods, on the other hand, deep learning-based methods such as auto-encoders, diffusion models, auto-regressive methods, and many more were adapted from the tabular domain to the graph domain.

First, Barabási–Albert model \citep{albert2002statistical} was proposed to capture the scale-free property observed in numerous real-world graphs. This property states that the degree distribution follows a power-law. The Barabási–Albert model has two parameters: the number of nodes and the number of edges to be added at each iteration. The graph is initialized with a fixed number of connected nodes. At each iteration, a new node is added and is connected to the existing nodes, with probability proportional to the current degree. Then, \citep{chen2007watts} introduced a model to deal with the small-world property, namely, the characteristics of high network clustering and short characteristic path length. The model consists of a regular lattice, typically a one-dimensional lattice with almost periodic boundary conditions. In other words, each vertex is connected to almost a fixed number of vertices nearest to it, and a small number of ‘shortcut’ bonds are added between randomly chosen vertices. BTER \citep{bter}, exploits the same concepts as the well-known Erdős–Rényi generation technique \citep{erdos-renyi} but in a two-level way, first modelling communities and then linking them together. Another statistical method is DANCer \citep{DANCer}, which creates a complete graph using preferential attachment \citep{barabasi-albert} and then performs micro (edge) and macro (community) updates so that the final graph matches the distribution of a reference. While these statistical techniques leverage important properties of large graphs, we believe they lack the expressiveness of deep models and they do not generate node attributes. 

On the other hand, deep learning models were proposed to learn graph generative models, the following paragraphs classify them in different families. 

In the Auto-Encoder (AE) family, the first Graph Variational AE (GVAE) \citep{graphvae} offered to generate a graph by sampling independent node representations from a known latent distribution and decoding it into a graph.
Some other approaches built upon this model achieved better graph quality, for example by extending the loss with higher level constraints on the graph \citep{gvaemm}. However, they all suffer from having to store a dense adjacency matrix, at least at generation time, which scales quadratically with the number of nodes, making them unscalable. 

More recently, many works have been released on performing graph generation with diffusion methods: NVDiff \citep{nvdiff}, GDSS \citep{gdss}, EDP-GNN \citep{edpgnn} and DiGress \citep{digress}. 
These models learn a reversible process from a graph representation to a known distribution, however, these methods too suffer from the need to store the dense adjacency matrix, both at train time and at generation time, making them unscalable.

There also exist SGGs based on reinforcement learning \citep{MNCE-RL}, adversarial networks \citep{molgan} or flow \citep{GrAD}. However, none of those works is currently considered state-of-the-art for large graph generation \citep{dggsurvey}. In addition, their application domain is limited to molecular graph generation.

Another family of SGGs is auto-regressive (AR) models, such as GraphRNN \citep{graphrnn}.
These embed each node in a recursive manner, and in doing so they update a state vector to condition the generation of a step.
Some of those models, like GRAN \citep{gran}, have been extended with attention layers for more expressiveness.
These models are very efficient in modelling small graphs as they do not suffer from the independent generation (of nodes/edges) of one-shot generation methods.
However, they often fail to represent high-level characteristics in the generated graphs as long-term dependencies are difficult to capture by recurrent models.

Some works enable recurrent models to accurately represent large graphs.
GraphGen \citep{graphgen} represents graphs by their minimum DFS codes\footnote{A graph (and its isomorphisms) can be uniquely identified by its minimum DFS code, without the need for an arbitrary ordering of nodes or edges.}.
This drastically reduces the size of the input space to the model.
BiGG \citep{bigg} is an auto-regressive model based on GraphRNN \citep{graphrnn} that represents the recursive process by binary trees, which reduces the number of recursive steps.
They also claim to scale with $\mathcal{O}(\sqrt{M \mathrm{log} N})$ memory-wise, $M$ being the number of edges in the graph.
However, neither GraphGen nor BiGG is able to generate node features in their original formulation\footnote{GraphGen is able to generate node and edge labels but not feature vectors.}.

Few works in the literature focused on random walks to learn generative models. They have the advantage of their invariance under node reordering. Additionally, random walks only include the nonzero entries of the adjacency matrix, thus efficiently exploiting the sparsity of real-world graphs. \citep{bojchevski2018netgan} proposed NetGAN, they train a generator for random walks, and a discriminator for synthetic and real random walks. After training, the generator is used to sample a paramount of random walks and a count matrix is computed for all edges. Then a threshold strategy is used to binarize this matrix.

Finally, there are works that combine hierarchical graph structure and deep models to tackle scaling issues of graphs while preserving good expressiveness.
One such model applies this hierarchical idea with chemistry motifs \citep{hierarchical-motifs} for molecule generation.
Similarly, but not restricted to molecules, HiGen \citep{higen} proposes an AR-Based method to exploit graphs' hierarchical structure.
It generates a high-level graph of communities in the first stage, then it extends each node into a community and each edge into inter-community links with a recursive model, potentially multiple times if there are more than two levels. However, they condition the expansion in the second stage only on the representation of the previous level and not on what has already been expanded elsewhere in the graph nor do they show the quality of the attribute generation.
Lastly, GELLCELL \citep{gellcell} proposes a technique for generating each community with the CELL model \citep{cell-netgan} and then connecting those communities by using a link prediction model based on XGBoost. 
Unfortunately, CELL \citep{cell-netgan} is based on statistical measures and was not shown to match the state of the art, and the linking of the communities is agnostic of the context around the nodes. 


Our work is the first to extend the BTER principle of two-step, top-down graph generation using deep networks, combining the efficiency of one-shot models (by means of model architecture choices) and the precision of conditional generation thanks to the refinement process. Our meta-algorithm is community-generator agnostic, unlike existing approaches in the literature. We show that the graphs generated using our method show statistical similarity in terms of topology and node features, while also leading to low privacy risks.

\section{Our Model}
\label{sec:SANGEA-model}

This section presents our proposed model for large-scale synthetic graph generation. The essence of our method is described in Section \ref{sec:SANGEA-algo-words}, then more details about the training and the generation procedures are given respectively in Section \ref{sec: SANGEA training process} and Section \ref{sec: SANGEA generation process}.


\subsection{The SANGEA Algorithm}
\label{sec:SANGEA-algo-words}
We propose a divide-and-conquer strategy for generating graphs. The main idea is to separate the graph into different communities of controllable size.
Usually, a graph is more densely connected within these communities than outside of them.
Each community is used to train one SGG model.
The SGG models are trained independently.
Once trained, they are used to generate a synthetic version of their respective community.
Then, the synthetic communities are patched together using a link prediction model.
Finally, we refine the synthetic graph's links until we are satisfied with the quality of the generated graph.
The pseudo-codes of SANGEA's training and generation steps are reported in Algorithms \ref{algo: SANGEA training} and \ref{algo: SANGEA generation}, respectively.
A detailed explanation of these algorithms follows in sections \ref{sec: SANGEA training process} and \ref{sec: SANGEA generation process}.

With this approach, we limit SGGs to graphs with fewer nodes, namely the communities.
Then, we use link prediction models to link the generated communities 
as these models are usually more lightweight at training and inference time than SGGs.
Finally, in the refinement step, we use extra link prediction models (refiners) to refine the final synthetic graph's topology.
The refiners are link prediction models that can be trained on $k$-hop neighbourhoods, rather than on a full graph, similarly to recursive models.
Besides, the SGGs are trained on communities.
Thus, at no point, does the full graph's \textit{dense} adjacency matrix need to be stored in memory when training SANGEA. Indeed, thanks to the community structure of the generation, we limit the memory cost of inference (generation) to the square of the size of the largest community rather than that of the full graph as further explained in the memory section. During training, the \textit{sparse} representation of the graph is sufficient to perform all operations. 



\RestyleAlgo{ruled}   
\SetKwComment{Comment}{}{}

\begin{algorithm}[!ht]
    \scriptsize
    \LinesNumbered
    \DontPrintSemicolon
    \SetKwInput{KwNotation}{Notation}
    \SetKwInput{KwCodeComment}{Comment}
    \KwNotation{
    $G[c_i \neq c_j]$: inter-community edges of $G$.
    $G[\bm{c} = k]$: node subgraph of $G$, only keeping nodes whose community is $k$.}
    \KwIn{A large graph $G$}
    \KwOut{A set of trained community generators, a base linker and a set of $k$-refiners}
    $\bm{c} \gets assign\_communities(G)$\Comment*[r]{Phase 1}\label{line: phase 1}
    $C \gets |unique(\bm{c})|$\Comment*[r]{}
    \For{$k \in [1, \ldots, C]$}{
        $g_k \gets G[\textbf{c} = k]$ \label{line: node filtering}\Comment*[r]{}
        $community\_generators[k] \gets train\_generator(g_k)$ \Comment*[r]{Phase 2}\label{line: phase 2}
    }
    $base\_linker \gets train\_autoencoder\left(\bigcup\limits_{1 \leq k \leq C} g_k\;,\;G[c_i \neq c_j]\right)$ \Comment*[r]{Phase 3}\label{line: phase 3}
    $base\_refiner \gets train\_autoencoder\left(G, G\right)$ \Comment*[r]{Phase 4}\label{line: phase 4}
    \For{$k \in [1, \ldots, C]$}{
        $k\_refiners[k] \gets finetune\_autoencoder(base\_refiner, G, g_k)$ \Comment*[r]{Phase 5 (a-b)}\label{line: phase 5ab}
    }
    $k\_refiners["inter"] \gets finetune\_autoencoder(base\_refiner, G, G[c_i \neq c_j])$ \Comment*[r]{Phase 5 (c)}\label{line: phase 5c}
    \KwCodeComment{Last argument of $train\_autoencoder$ and $finetune\_autoencoder$ are the edges used as labels in the loss.}
    \caption{SANGEA learning process}
    \label{algo: SANGEA training}
\end{algorithm}

\subsection{Training Process}\label{sec: SANGEA training process}
\begin{figure}[!ht]
    \centering
    \includegraphics[width=\textwidth]{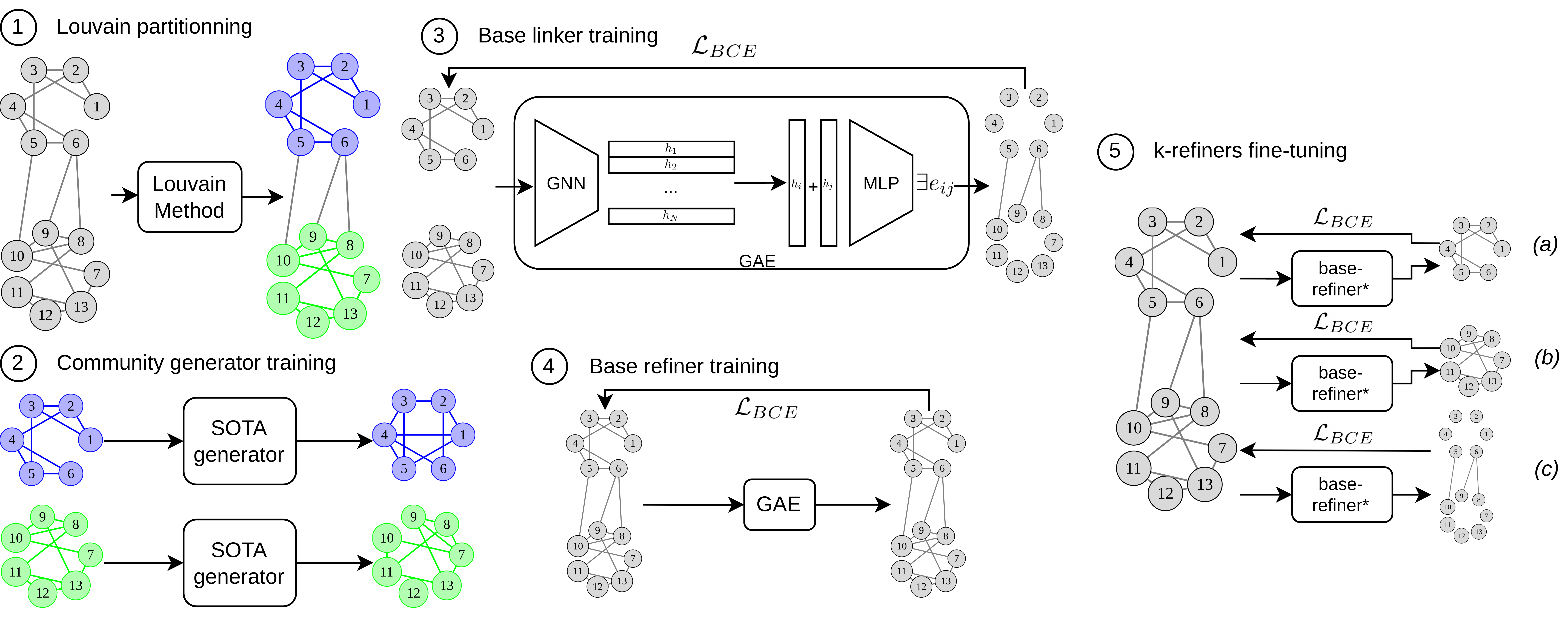}
    \caption{Full training procedure of the SANGEA generation method}
    \label{fig: full SANGEA training procedure}
\end{figure}

Algorithm \ref{algo: SANGEA training} shows the entire training process of the SANGEA algorithm. This process is also depicted in Figure \ref{fig: full SANGEA training procedure}. The training is divided into 5 phases.


\textit{i) Louvain Partitioning:} line \ref{line: phase 1} in Algorithm \ref{algo: SANGEA training} shows a call to $assign\_communities$. This function will assign to each node a label, i.e. a community, as given by the Louvain method \citep{louvain}. This greedy algorithm, designed for very large graphs, aims at optimizing modularity, which measures how densely connected the communities are within themselves and how sparse the links between different communities are. 
\textit{ii) Community Generator Training:} once the communities have been found, the original, large graph can be separated into independent, disconnected components that correspond to the graphs defined by the nodes of each community. 
These are graphs of smaller size than the original graph, thus any generation technique that only applies to small graphs can be trained on them.
In this phase, we train one generator per community, with each generator being totally independent of the others.
Note that the rest of the SANGEA algorithm is agnostic of what method is used to generate the communities. 

\textit{iii) Base Linker Training:} the base linker is a graph autoencoder (GAE) model composed of a GNN encoder module and a MLP decoder module.
This model is trained for the link prediction task.
However, no message passing \citep{gilmer2017neural} over inter-community links is allowed at this stage, thus message passing is allowed only over the intra-community links, and the model is trained to predict the inter-community links only.

\textit{iv) Base Refiner Training: }in Phase 4, we create a new GAE, possibly with different hyperparameters than the base linker, also trained for the link prediction task.
However, the message passing now goes through the whole training graph and the edges used as training samples in the loss are also all the edges of the original training graph. 
This model is never used for prediction.
However, it is used in Phase 5 of the training described below.

\textit{v) k-Refiners Fine-Tuning: }in this phase, the base refiner learned in Phase 4 is copied and then further trained using the entire original graph for message passing and the creation of embeddings. However, only a specific subset of links are used as samples in the loss. Specifically, we create $C+1$ copies of the base refiner: one for the links within each community and one for the inter-community links.
This fine-tune approach has two main goals: (i) It is a reasonable assumption that what is learned on the whole graph is transferable to specific parts of that graph, especially if the model is fine-tuned on that part of the graph; 
(ii) Some communities can be tiny, and training a model on a few samples without overfitting is a complicated task. Using this base refiner/fine-tuning approach, we still obtain good generalization results for those communities.


\subsection{Generation Process}\label{sec: SANGEA generation process} 

Once the training has been completed, it is time to generate a large graph using these trained models. 

\begin{algorithm}[!ht]
    \scriptsize
    \setcounter{AlgoLine}{0}
    \LinesNumbered
    \DontPrintSemicolon
    \SetKwInput{KwNotation}{Notation}
    \KwNotation{$\hat{G}[c_i \neq c_j]$: inter-community edges of $\hat{G}$.
    $\hat{G}[(c_i = label) \land (c_j = label)]$: set of links of $\hat{G}$ where both end nodes are within community $label$.
    $K_{label}, K_{inter}$: set of all possible edges that match that label (i.e. all possible inter-community edges or all possible edges within a community).}
    \KwIn{The models trained at the training phase, $C$ the number of communities, $R$ the number of refinement steps.}
    \KwOut{Generated graph $\hat{G}$.}
    \For{$k \in [1, \ldots, C]$}{ \label{line: start gen phase 1}
        $\hat{g}_k \gets community\_generators[k].generate()$\Comment*[r]{}
    }
     $\hat{G} \gets \bigcup\limits_{1 \leq k \leq C} \hat{g}_k$\Comment*[r]{}
     $\textbf{s} \gets base\_linker.score\_links(\hat{G}, K_{inter})$\Comment*[r]{}
     $\hat{G} \gets \hat{G} \bigcup sample(\textbf{s})$\label{line: end gen phase 1}\Comment*[r]{}

    \For{$r \in [1, \ldots, R]$}{\label{line: start gen phase 2}
        \For{$label \in [1, \ldots, C, "inter"]$}{
            \eIf{$label$ is $"inter"$}{
                 $\Tilde{\textbf{s}} \gets \textbf{1} - k\_refiners["inter"].score\_links(\hat{G}, \hat{G}[c_i \neq c_j])$\Comment*[r]{}
            }{ 
                 $\Tilde{\textbf{s}} \gets \textbf{1} - k\_refiners[label].score\_links(\hat{G}, \hat{G}[(c_i = label) \land (c_j = label)])$\Comment*[r]{}
            }
             $\textbf{s} \gets k\_refiners[label].score\_links(\hat{G}, K_{label})$\Comment*[r]{}

             $\hat{G} \gets \hat{G} \setminus sample(\Tilde{\textbf{s}})$\Comment*[r]{}
             $\hat{G} \gets \hat{G} \cup sample(\textbf{s})$\Comment*[r]{}
        }
    }\label{line: end gen phase 2}
    \caption{SANGEA generation process}
    \label{algo: SANGEA generation}
\end{algorithm}


\textit{i) Base Generation:} graph generation using SANGEA happens in two phases.
In this first phase, for each community, a graph is generated using the corresponding community generator, resulting in a collection of synthetic graphs, one per community.
This collection of disconnected components is then used for the message-passing of the base linker and the inter-community edges are predicted in one shot.
We generate as many edges as there were in the original graph.

\textit{ii) Refinement: }due to the independence of the community generators, and due to the base linker's lack of access to the whole graph (in terms of message passing), we designed a refinement phase where we iteratively update the graph by means of a new link predictor that, this time, has access to all links for the message passing.
Therefore, at each refinement step, we will input the full graph to all the $k$-refiners, each updating a different part of the graph. 
Doing this, we condition the updates of the links on the current state of the graph, in a way that is analogous to recurrent models. 
However, this is all done using one-shot models.
We can perform this phase $R$ times for the desired amount of refinements steps.
Each refinement step replaces edges that have low scores with ones that have high scores, with the objective of improving the final topology of the generated graph.
The number of refinements controls the trade-off between privacy and generation quality, and would in fact depend on the actual downstream use of the generated data. 

\subsection{Memory Usage}
    
\textit{i) At Training Time: }most one-shot generation techniques usually require the model to store in memory a dense adjacency matrix. 
This causes these models not to scale very well. Here, we show the theoretical memory upper-bound usage of SANGEA's full procedure.
Let us imagine we have a large graph of $N$ nodes and $M$ edges. 
The Louvain method, which runs with $\mathcal{O}(M)$ memory cost, yields communities for the large graph, the largest of which has $N_{c^*}$ nodes, with $c^*$ being the biggest community.
Because we control the size of the communities, we can assume that $N_{c^*} \ll N$ \citep{multiscaleresolutionlouvain}. 
In the worst case, the method used as a community generator saves the whole dense adjacency matrix, which would imply a memory consumption proportional to $\mathcal{O}(N_{c^*}^2)$. 
For the base linker, the full training graph goes through the GNN layers that store, for each node, one latent representation.
This means that its memory impact is proportional to $N$, and then, for each edge used for the loss, the pair of corresponding node representations go through an MLP, which only needs to store gradients per node, having a memory cost proportional to $N$ as well. 
The memory cost can be further reduced.
In order to compute a node embedding, one may store the $k$-hop neighbourhood of that node.
Because loss samples are edges, to perform a backpropagation, we only need the $k$-hop neighbourhoods of the two end nodes of that edge.
This makes the memory impact in $\mathcal{O}(N_k)$ with $N_k$ being the maximum number of nodes in all $k$-hop neighbourhoods. 
Assuming sparsity, $N_k$ is often much lower than $N$. This memory frugality comes at a computational expense.
In practice, the value of $k$ can be parameterized to match the memory capacity of the device running the computation.
With this method, the whole memory impact of the base-linker at training time is measured in $\mathcal{O}(N_k)$.
This holds for all other GAEs of the training process.
This shows that at training time, our models can run in memory complexity bounded by $\mathcal{O}(\max(N_{c^*}^2, M))$, omitting $N_k$ as it can be assumed to be smaller than $M$.


\textit{ii) At Inference time: }in the worst case, each community generator requires to generate a dense adjacency matrix for each community. This means that the memory impact is bounded in $\mathcal{O}(N_{c^*}^2)$. Then comes the base linker. This model scores all possible edges not within communities. The number of such edges is $N_{inter} = N^2 - \sum_{i = 1}^{C} N_i ^2$, which grows in $\mathcal{O}(N^2)$. However, rather than scoring all edges at once and then sampling, it is possible to score all edges between a pair of communities, sample amongst those, and then discard the memory used for those scores. This means that at a given time, we only store all the possible edges between a pair of communities, thus making the generation bounded in memory by $\mathcal{O}(N_{c^*}^2)$.

Applying the same reasoning to refinement, for each community-refiner, we will be bounded in memory by $\mathcal{O}(N_i^2)$, and for the inter-refiner, using the same trick as for the base linker, we are bounded in $\mathcal{O}(N_{c^*}^2)$. Combining inference memory complexities, we obtain a final memory upper-bound growing in $\mathcal{O}(N_{c^*}^2)$ for the whole process.





\section{Experiments}
\label{sec:exps}
In the present work, we present an approach to scale up any SGG to large graphs.
In this Section, we present the experiments that we designed to validate that our method is indeed efficient and generates high-quality samples.
This section aims at answering the following research questions (\textbf{RQs}):

\begin{enumerate}
    \item Which models in the literature handle large graphs? 
    \item Can we make state-of-the-art models scalable to large graphs thanks to our approach?
    \item Is one approach better than the other in terms of utility and privacy? 
    \item Does our approach bring performance gains compared to state-of-the-art approaches that deal with large graphs?
\end{enumerate}


\begin{table}
\footnotesize
\begin{center}
    \begin{tabular}{l c c c c} 
    \toprule
     & \multicolumn{4}{c}{\textit{Max. conn. comp. properties}} \\
    \cline{2-5}
     & Nodes & Edges & Features & Classes\\
    \midrule
    Cora & 2485 & 5069 & 1433 & 7\\ 
    CiteSeer & 2120 & 3679 & 3703 & 6 \\ 
    IMDB & 10384 & 16097 & 3066 & 4 \\
    Amazon Computers & 13381 & 245778 & 767 & 10 \\
    Flickr & 89250 & 449878 & 500 & 7 \\
    \bottomrule
\end{tabular}
\end{center}
\caption{Summary of the datasets (\textit{Maximum Connected Component}).}
\label{tab:datasets}
\end{table}

\subsection{Data Description}

Table \ref{tab:datasets} lists the various datasets used for our empirical evaluation and statistics considering their \textit{maximum connected component}.
We chose one-graph datasets from \cite{FeyLenssen2019}.
Cora and CiteSeer are citation networks, IMDB's nodes represent movies, actors, or directors. Amazon\footnote{In further developments we may refer to the Amazon Computers dataset simply as the Amazon dataset.}'s nodes represent products, and edges represent the co-purchased relations of products. The Flickr dataset is an ensemble of images presented in a graph where the nodes are the images, and the edges are assigned based on the presence of some shared properties (e.g., identical geographical area, identical exhibition space, remarks made by the identical individual). 

\subsection{Evaluation Metrics}
\label{sec:eval_metrics}
Multiple aspects have been considered to compare the proposed approaches. First is the structural and attribute similarity between the generated and original graphs \citep{thompson2022evaluation}.
Second, the utility of the generated graphs for downstream tasks.
Third, training, generation time, and memory consumption in order to assess the ability to handle large graphs. Finally, the privacy risk associated with the generated graph.

\paragraph{}
Multiple metrics to assess the structural similarity between the generated and original graphs have been considered in our experiments.
Namely, degree histogram, degree centrality, closeness centrality, eigenvector centrality, and clustering coefficient.
The graphs are represented as normalized versions of these metrics and then compared using the Wasserstein distance \citep{wasserstein}. 
Compared to the widely used Maximum Mean Discrepancy (MMD), the Wasserstein distance is more reliable.
Indeed, the MMD requires additional parameters to be chosen, and issues of sensitivity to these choices have been recently raised by \citet{o2021evaluation}. 
Node attribute similarity is implicitly taken into account by a distance measure over node embedding distributions, specifically because node embeddings also depend on node features. Nevertheless,
for this comparison, we opted for the MMD, as it does not require the complex tasks of binning and then computing optimal transport between histograms in a multidimensional space. Two versions of a Graph Convolutional Network (GCN), one untrained and one trained on the link prediction task, are used to embed the input graphs into an embedding space of size 16.
Then the MMD is used to compute the distance between the embeddings of the generated graph and the ones associated to the original graph. 
We also assess graph utility by training a GNN model on the generated graphs, on the link prediction task, and testing on the original graph.
In fact, we train a VGAE link predictor on the generated graph and measure AUROC on the original graph. 


Since our use-case consists in training a SGG from a single training graph, we evaluate the privacy concerns that it may imply.
We choose to evaluate privacy using the Nearest Neighbour Distance Ratio (NNDR) on node embeddings of the original and generated graphs. This metric is popular in the privacy domain \citep{gussenbauer2021ai}.
The full methodology works as follows: we first train a GCN embedder (details in the supplementary material) on the original data on the node classification task. 
Then, for each node embedding in the generated set, we compute the distance to all nodes in the original set using the Euclidean distance.
Thus, we have a distance vector $\bm{d}^{i} \in \mathbb{R}^N$, for the $i$-th node in the generated graph, with $N$ being the number of nodes of the original graph. 
Finally, if $d_1^i$ and $d_2^i$ are respectively the smallest and second smallest distances in $\bm{d}^i$, the NNDR for node $i$ can be computed as: $ NNDR_i = \frac{d_1^i}{d_2^i}$.
For each generated node, NNDR measures the ratio between its two closest neighbours in the training set.
It can be interpreted as \textit{the higher the ratio, the harder it will be to infer that a given target node was a member of the SGG's training set}.
Since this metric is dependent on the embedder chosen, we do the following. We estimated the NNDR between the original graph and itself, then between the original graph and perturbed versions of itself, with increasing perturbation strength.
Then, we chose the embedder that shows (i) a low NNDR value between the original graph and itself, and (ii) an increasing NNDR value on increasingly perturbed versions of the original data.

\subsection{Experimental Protocol}



\paragraph{}
All experiments are performed on a machine running Intel(R) Xeon(R) Gold 6134 \\CPU@3.20GHz processor with 32 physical cores, with 1 Nvidia Tesla V100 card with 32GB GPU memory, and 128GB RAM with Ubuntu 18.04.6 operating system. 

\paragraph{}
The first step in our experiments is to assess the capability of these models to deal with large graphs. Five approaches  \citep{bigg, gdss, nvdiff, graphgen, gvaemm} have been considered in our experiments.\footnote{Our most direct competitors are HiGen \citep{higen} and GELLCELL \citep{gellcell} but neither of them has code publicly available nor do they report results on large, attributed real-world graphs. We, therefore, consider GraphGen and BiGG as our closest scalable competitors.} Based on the results of this step, competitors to our approach will be identified according to their ability to handle big graphs. 
Then, these models will be used as community generators within our proposed approach.
First, communities are identified using the Louvain algorithm \citep{louvain}.
They are used as training examples for the community generators. Each community generator is trained on one subgraph (i.e. community), and it is done for all considered state-of-the-art approaches. 
Once communities are generated, our proposed approach is used to generate the final version of the graph (more details in Section \ref{sec:SANGEA-model}). The next step is to compare the different variants of our proposed approach on multiple dimensions: statistical properties, utility metrics, scalability, and privacy risk. In addition, a comparison to the selected state-of-the-art approaches will be performed.  In our experiments we have chosen to set the number of refinements $R=30$, the model parameters for all experiments are reported in Table 13 in the supplementary material. We search through all of those values through hyper-parameter optimization using the Optuna framework for the downstream task of link prediction.
For the MMD metric, we used a Gaussian RBF with parameter sigma = 0.5, and for the community partitioning, we used a resolution parameter of 1 for the Cora and Citeseer datasets, to ensure sufficiently large communities, and of 1.5, 5.5 and 5.5 for the IMDB, Amazon and Flickr datasets respectively, to have communities of around a thousand nodes maximum.

\subsection{Results and Analysis}
In this section, detailed answers to the questions raised in the introduction of Section \ref{sec:exps} are given, supported by numerical results.

\textit{Scalability:} In Table \ref{tab:topo_res_imdb_amazon}, one can notice that three out of five tested approaches fail to train the generative models on IMDB's maximum connected graph. Only GraphGen and BiGG are capable of dealing with this graph.  In Table \ref{tab:topo_res_imdb_amazon}, it is clear that all implemented state-of-the-art approaches fail to train on Amazon Computers's graph, which has 15 times more edges than IMDB. We refer the reader to the supplementary material for a similar result on the Flickr dataset. We note that Flickr dataset is very large, and has been used only to provide scalability results; the communities were copied from the original graph and were linked together using our proposed approach. We conclude from these results that we have only two competitors for medium-sized graphs. Furthermore, it is not possible to train current state-of-the-art approaches on big graphs, given our computational capacity and the limitations of existing models from the literature. For the running time metrics of state-of-the-art approaches and our model, we refer the interested reader to the appendix.

\begin{table}[!ht]
\scriptsize
    \begin{center}
    \setlength{\tabcolsep}{6pt}

    \begin{tabular}{l ccccc c ccccc}
        \toprule
        & \multicolumn{5}{c}{IMDB Dataset} & & \multicolumn{4}{c}{Amazon Computers Dataset} \\
        \cline{2-6} \cline{8-11}
        & \multicolumn{2}{c}{Others} & & \multicolumn{2}{c}{Ours} & & \multicolumn{1}{c}{Others} & & \multicolumn{2}{c}{Ours} \\
        \cline{2-3} \cline{5-6} \cline{8-8} \cline{10-11}
         & \multirow{2}{*}{GraphGen} & \multirow{2}{*}{BiGG} &  & Sa & Sa & & \multirow{2}{*}{All} & & Sa & Sa \\
         & &  &  & (GDSS) & (NVDiff) & &  & & (GDSS) & (NVDiff)\\
        \midrule
        MMD (tr.) & - & - & & \textbf{0.305} & 0.379 &  & \multirow{9}{*}{TO/OOM} & &  0.419 & \textbf{0.375} \\
        MMD (untr.) & - & - &  & \textbf{0.00210} &  0.0713 & &  &  & 0.0426 & \textbf{0.0329} \\
        WS spectral & \textbf{1.53e-3} & 9.88e-3  & & 5.66e-3 & 3.50e-3 & &  &  & \textbf{1.68e-3} & 1.74e-3  \\
        WS deg. hist. &  125e-4 & 29.7e-4  & &  \textbf{8.58e-4} & 18.2e-4 & &  &  & \textbf{8.56e-5} &  11.0e-5\\
        WS deg. cent. & 3.19e-5 & 1.11e-5 & & \textbf{1.05e-5} & 5.40e-5 &   &  &  & \textbf{2.04e-6} &  9.88e-6 \\
        WS clos. cent. & 26.34e-5 & 8.87e-6 & & \textbf{2.84e-6} & 10.0e-6 &   &  &  &\textbf{2.76e-6}  & 12.8e-6 \\
        WS eig. cent. & \textbf{1.62e-5}& 5.07e-5 & & 4.12e-5 & 3.59e-5 &  &  &  &\textbf{1.78e-5} & 3.33e-5   \\
        WS clust. coeff. & 29.7e-6  & \textbf{1.44e-6}  & & 15.1e-6 & 20.7e-6 & &  &  &\textbf{4.24e-5}  & 5.78e-5 \\
        AUROC (LP)  & 0.74 & 0.73 & &  \textbf{0.76} & 0.74 &  &  &  & 0.814 & \textbf{0.833} \\
        \bottomrule
    \end{tabular}
    \end{center}
    \vspace{-8pt}
    \caption{ Structural and attribute similarity results on IMDB and Amazon Computers datasets.  OOM stands for Out Of Memory and TO stands for TimeOut. Any competitor among GraphGen, BiGG, NVDiff, GVAEmm or GDSS not shown in the table indicates an OOM or TO. Sa(GDSS) stands for SANGEA using GDSS as community generator.}
    \label{tab:topo_res_imdb_amazon}
\end{table}

SANGEA is able to generate graphs with sizes equivalent to those of Amazon Computers' size, as well as Flickr' size, which has over 6 times the number of nodes and twice the edges.
All state-of-the-art approaches considered here, however, fail to do so. (running times available in the supplementary material). While the memory resource usage is greatly improved, all the independent steps of SANGEA add a consequent time overhead. This overhead can be greatly reduced with parallel training, which is facilitated by SANGEA's innate design.
This can be explained by the fact that, no matter the size of the graph, we scale with the size of the largest community, which we control, thus enabling our model to choose where we set the time-memory cost trade-off.

\paragraph{}
\textit{Graph generation quality:} Table \ref{tab:topo_res_imdb_amazon} reports results on the structural and attribute similarity between the original training graph and the generated one on the IMDB and Amazon Computers datasets \citep{thompson2022evaluation}.
SANGEA allows for node feature generation via the community generators. This is the case for Sa(GDSS) and Sa(NVDiff), while other scalable models do not generate node features. Our method shows to be superior on the MMD over GCN embedding metrics. On the IMDB dataset, Sa(GDSS) closely follows GraphGen when it does not outperform it on most statistical topology metrics and outperforms BiGG on most of them. Increased performance on the downstream task of Link Prediction for the IMDB dataset might again be explained by the lack of node feature generation in the competitors' models. As to the Amazon Computers dataset, our models show superior performance since no other model could scale to that dataset size. These results suggest that on medium to large graphs, SANGEA can match, and even surpass other state-of-the-art methods.

\paragraph{}
\textit{Privacy:} Table \ref{tab: privacy} shows the NNDR values obtained on 4 datasets, using SANGEA and also two mode models from the state of the art, NVDiff and GDSSS, for which the results are reported on generated communities only (because we were not able to generate the full graph with these models). We compare the graphs generated with three baselines, the original graph and two perturbed versions of that graph; one at 50\% and one at 75\%. Perturbation at $p\%$ corresponds to the original data where $p\%$ of the edges have been replaced by random ones, and $p\%$ of the node feature matrix has been changed. The table shows that, for both of our generated graphs and for each dataset, they always at least match the 50\% perturbation, often reaching the level of, or surpassing the 75\% perturbation. This shows that even though our method generates graphs of high utility and close statistical properties, it achieves the privacy level, for the individual nodes of the training graph, of at least a 50\% perturbation graph, which is higher than the privacy levels (NNDR value) reached by using NVDiff or GDSS alone.
\begin{table}
    \scriptsize
    \centering
    \begin{tabular}{lcccc}
        \toprule
         & \multicolumn{4}{c}{Datasets}\\
        \cline{2-5}
         & CiteSeer & Cora  & IMDB & Amazon \\
        \midrule
        Orig. data & 0.05 ± 0.21  & 0.05 ± 0.20 &  0.32 ± 0.43 & 0.50 ± 0.43 \\
        Pert. data (50\%) & 0.89 ± 0.01 & 0.85 ± 0.14 &  0.81 ± 0.25 & 0.89 ± 0.14\\
        Pert. data. (75\%) & 0.90 ± 0.09  & 0.90 ± 0.09 &  0.97 ± 0.10 & 0.91 ± 0.10\\
        \midrule
        NVDiff & 0.66 ± 0.39& \textbf{0.98 ± 0.03} & \textbf{0.99 ± 0.06} & 0.88 ± 0.25 \\
        GDSS & \textbf{0.84 ± 0.00} & 0.69 ± 0.00 & 0.22 ± 0.00 & \textbf{0.92 ± 0.01}\\
        \midrule
        Sa(NVDiff) & 0.90 ± 0.09 & 0.86 ± 0.16 & 0.92 ± 0.15 & 0.91 ± 0.20 \\
        Sa(GDSS) & \textbf{0.91 ± 0.09} & \textbf{0.99 ± 0.03} & \textbf{0.97 ± 0.11} & \textbf{0.99 ± 0.01}\\
        \bottomrule
    \end{tabular}
    \caption{NNDR (mean ± stand. dev.) of generated graphs. NVDiff and GDSS without SANGEA were evaluated on the largest community of each dataset, all others on the full dataset.}
    \label{tab: privacy}
\end{table}

\begin{table}
\scriptsize
    \centering
    \begin{tabular}{lccccc}
        \toprule
         & \multicolumn{2}{c}{Cora, Sa(NVDiff)} & & \multicolumn{2}{c}{IMDB, Sa(GDSS)} \\
        \cline{2-3}  \cline{5-6}
         & \textit{w.o. ref.} & \textit{w. ref.} &  & \textit{w.o. ref.} & \textit{w. ref.}\\ 
         \midrule
        MMD (gcn tr.) & 13.7e-2 & \textbf{9.6e-2} &  & 0.438 & \textbf{0.305} \\
        MMD (gcn untr.) & 17.9e-1 & \textbf{4.9e-1} &  & 0.0631 & \textbf{0.00210} \\
        WS spectral & 11.e-4  & \textbf{5.22e-4} & &  43.2e-3 &  \textbf{5.66e-3} \\
        WS deg. hist. & 17.6e-3 & \textbf{1.46e-3} &  & 41.4e-4 &  \textbf{8.58e-4}\\ 
        WS deg. cent. & 9.25e-05 & \textbf{6.66e-5} & & \textbf{1.05e-5} & \textbf{1.05e-5} \\
        WS clos. cent. & 10.6e-5 & \textbf{9.10e-5} &  & 26.2e-6  & \textbf{2.84e-6} \\
        WS eigenv. cent. & 13.1e-4 & \textbf{3.88e-4} &  & 31.2e-5 &  \textbf{4.12e-5} \\ 
        WS clust. coeff. & 35.8e-4 & \textbf{4.64e-4} &  & 45.6e-5  & \textbf{1.51e-5} \\ 
        AUROC (LP)  & 0.71 & \textbf{0.74} &  &  0.71 & \textbf{0.76} \\ 
        \bottomrule
    \end{tabular}
    \caption{Ablation study of the refinement process}
    \label{tab:ablation study}
\end{table}

\paragraph{}
\textit{Ablation study}: One of the main novelties of this work, which also differentiates it the most from HiGen \citep{higen} and GELLCELL \citep{gellcell}, is the refinement process.
This process conditions the predictions and updates of the links in the generated graph on previously generated links and nodes.
Table \ref{tab:ablation study} shows, for two different datasets, using two different community generators, that the refinement process improves the utility of the final generated graphs, which confirms that this process is a valuable feature of the method proposed in the present work. Results on more datasets are provided in the supplementary material, and highlight similar results.



    

\section{Conclusion}
\label{sec:conclusion}

We presented a novel approach called SANGEA, a lightweight method to scale graph generative models to many nodes and edges. It generates from a single large training graph another large graph that matches the statistical properties of the original one while achieving high privacy scores. Extensive experiments have been conducted to assess the effectiveness of our approach.
Five state-of-the-art approaches have been considered from the literature to benchmark against.
We show in our experiments that SANGEA can work with graphs with up to 90,000 nodes and 450,000 edges, while the chosen literature approaches fail. 
Moreover, the quality of generation has been assessed using multiple graph quality metrics.
Numerical results show a high similarity between our generated graph and the original one compared to our direct competitors. 
SANGEA also achieves a better utility score on the link prediction task.
In addition, because of the single large training graph constraint of our setting, a privacy assessment methodology has been proposed and discussed.
Our results show that the generated graphs naturally obtain high-privacy scores and hence is low-risk.





Our proposed approach suffers from a set of limitations. First, the feature generation is conditioned by the community generator offering this property. Moreover, our feature generation is limited to node features. Then, the input graphs are assumed to be static. In many applications, dynamic evolving graphs are of interest. For example, we can have user dynamics like join/leave events and evolving relationships in such data. These limitations are interesting directions to extend the capabilities of our approach in future work.


%
%
%

\bibliography{bibliography.bib}

\end{document}